\documentclass[10pt,twocolumn,letterpaper]{article}

\usepackage{cvpr}
\usepackage{times}
\usepackage{epsfig}
\usepackage{graphicx}
\usepackage{subfigure}
\usepackage{epstopdf}
\usepackage{amsmath}
\usepackage{amssymb}
\usepackage{dcolumn}

\usepackage[pagebackref=true,breaklinks=true,letterpaper=true,colorlinks,bookmarks=false]{hyperref}

\cvprfinalcopy 

\def\cvprPaperID{4675} 

\ifcvprfinal\fi
\begin{document}

\title{Look More Than Once: An Accurate Detector for Text of Arbitrary Shapes}
\author{Chengquan Zhang$^1$\thanks{Equal contribution. This work is done when Borong Liang is an intern at Baidu Inc.}  \quad Borong Liang$^{2*}$ \quad Zuming Huang$^{1*}$ \quad Mengyi En$^1$ \\
\quad Junyu Han$^1$ \quad Errui Ding$^1$ \quad Xinghao Ding$^2$\thanks{Corresponding author.}\\
Department of Computer Vision Technology (VIS), Baidu Inc.$^1$ \\
\quad Fujian Key Laboratory of Sensing and Computing for Smart City, Xiamen University$^2$\\
{\tt\small \{zhangchengquan,huangzuming,enmengyi,hanjunyu,dingerrui\}@baidu.com}\\
{\tt\small liangborong@stu.xmu.edu.cn dxh@xmu.edu.cn}
}
\maketitle

\begin{abstract}
Previous scene text detection methods have progressed substantially over the past years.
However, limited by the receptive field of CNNs and the simple representations like rectangle bounding box or quadrangle adopted to describe text, previous methods may fall short when dealing with more challenging text instances, such as extremely long text and arbitrarily shaped text. To address these two problems, we present a novel text detector namely LOMO, which localizes the text progressively for multiple times (or in other word, LOok More than Once). LOMO consists of a direct regressor (DR), an iterative refinement module (IRM) and a shape expression module (SEM). At first, text proposals in the form of quadrangle are generated by DR branch. Next, IRM progressively perceives the entire long text by iterative refinement based on the extracted feature blocks of preliminary proposals. Finally, a SEM is introduced to reconstruct more precise representation of irregular text by considering the geometry properties of text instance, including text region, text center line and border offsets. 
The state-of-the-art results on several public benchmarks including ICDAR2017-RCTW, SCUT-CTW1500, Total-Text, ICDAR2015 and ICDAR17-MLT confirm the striking robustness and effectiveness of LOMO. 
\end{abstract}

\section{Introduction}\label{sec:intro}
Scene text detection has drawn much attention in both academic communities and industries due to its ubiquitous real-world applications such as scene understanding, product search, and autonomous driving. Localizing text regions is the premise of any text reading system, and its quality will greatly affect the performance of text recognition.
Recently general object detection algorithms have achieved good performance along with the renaissance of CNNs. However, the specific properties of scene text, for instance, significant variations in color, scale, orientation, aspect ratio and shape, make it obviously different from general objects. Most of existing text detection methods~\cite{liao2017textboxes,liao2018rotation,ma2018arbitrary,zhou2017east,he2017deep} achieve good performance in a controlled environment where text instances have regular shapes and aspect ratios, e.g., the cases in ICDAR 2015~\cite{karatzas2015icdar}. 
Nevertheless, due to the limited receptive field size of CNNs and the text representation forms, these methods fail to detect more complex scene text, especially the extremely long text and arbitrarily shaped text in datasets such as ICDAR2017-RCTW~\cite{shi2017icdar2017}, SCUT-CTW1500~\cite{yuliang2017detecting}, Total-Text~\cite{ch2017total} and ICDAR2017-MLT~\cite{nayef2017icdar2017}.


\begin{figure}
\centering
\includegraphics[scale=0.5]{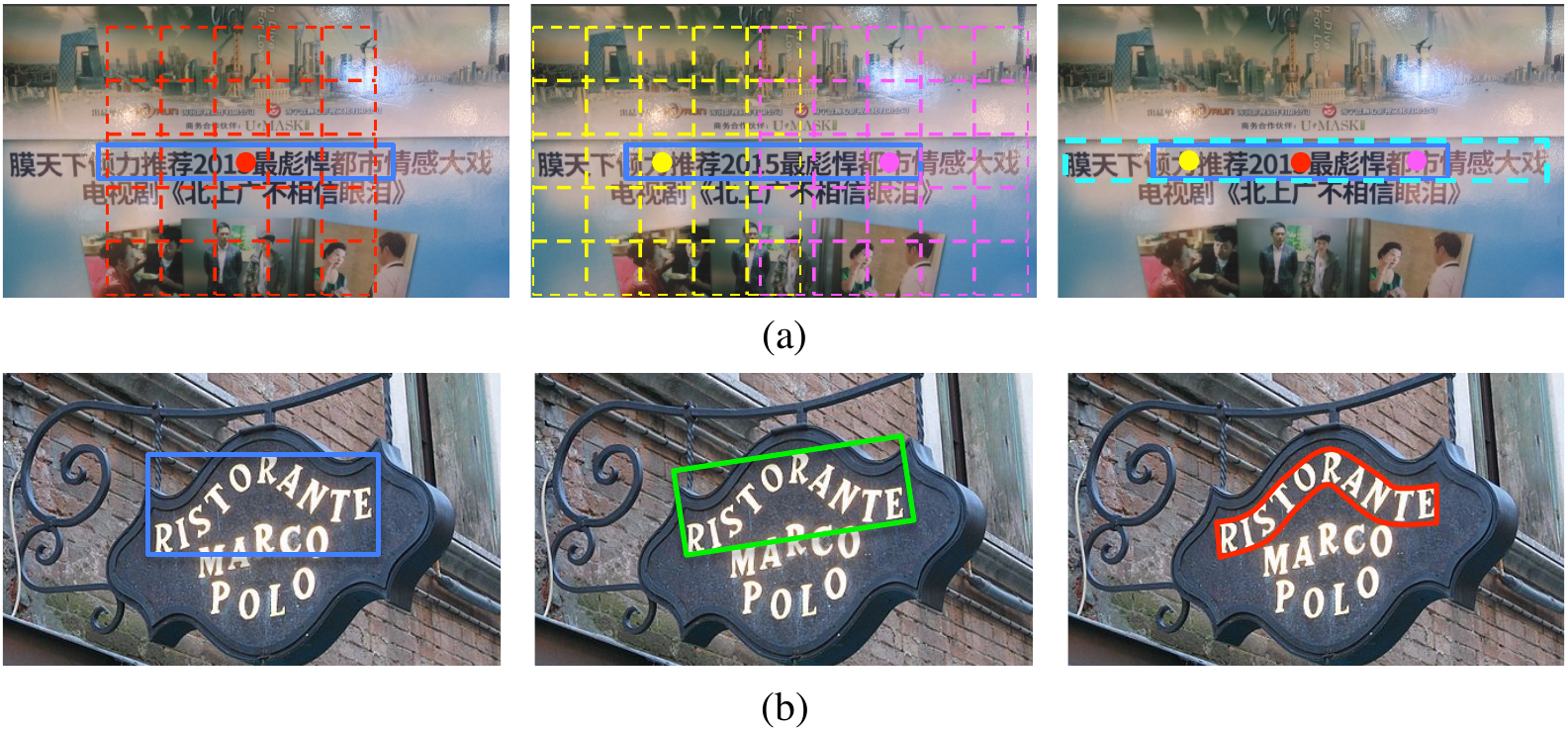}
\caption{Two challenges of text detection: (a) The limitation of receptive field size of CNN; (b) Comparison of different representations for text instances.}
\label{fig:intro}
\vspace{-0.5cm}
\end{figure}

When detecting extremely long text, previous text detection methods like EAST~\cite{zhou2017east} and Deep Regression~\cite{he2017deep} fail to provide a complete bounding box proposal as the blue box shown in Fig. \ref{fig:intro} $(a)$, since that the size of whole text instance is far beyond the receptive field size of text detectors.
CNNs fail to encode sufficient information to capture the long distant dependency. 
In Fig. \ref{fig:intro} $(a)$, the regions in grids mainly represent the receptive field of the central point with corresponding color. The blue quadrangle in Fig. \ref{fig:intro} $(a)$ represents the predicted box of mainstream one-shot text detectors~\cite{zhou2017east,he2017deep}. 
The mainstream methods force the detectors to localize text of different length with only once perception,  which is contrary to human visual system in which LOok More than Once (LOMO) is usually required.
As described in ~\cite{huang2016detecting}, for a long text instance, humans can only see a part at the first sight, and then LOok More than Once (LOMO) until they see the full line of text. 
 
In addition, most of existing methods adopted relatively simple representations (e.g., axis-aligned rectangles, rotated rectangles or quadrangles) for text instance, which may fall short when handling curved or wavy text as shown in Fig. \ref{fig:intro} $(b)$. Simple representations would cover much non-text area, which is unfavorable for subsequent text recognition in a whole OCR engine. A more flexible representation as shown in the right picture of Fig. \ref{fig:intro} $(b)$ for irregular text can significantly improve the quality of text detection.

In order to settle the two problems above, we introduce two modules namely iterative refinement module (IRM) and shape expression module (SEM) based on an improved one-shot text detector namely direct regressor (DR) which adopts the direct regression manner~\cite{he2017deep}. 
With IRM and SEM integrated, the proposed architecture of LOMO can be trained in an end-to-end fashion.
For long text instances, the DR generates text proposals firstly, then IRM refines the quadrangle proposals neatly close to ground truth by regressing the coordinate offsets once or more times. 
As shown in middle of Fig.~\ref{fig:intro} $(a)$, the receptive fields of yellow and pink points cover both left and right corner points of text instance respectively. 
Relying on position attention mechanism, IRM can be aware of these locations and refine the input proposal closer to the entire annotation, which is shown in the right picture of Fig.~\ref{fig:intro} $(a)$. The details of IRM is thoroughly explained in Sec.~\ref{sec:irm}.
For irregular text, the representation with four corner coordinates struggles with giving precise estimations of the geometry properties and may include large background area. Inspired by Mask R-CNN \cite{he2017mask} and TextSnake \cite{long2018textsnake}, SEM regresses the geometry attributes of text instances, i.e., text region, text center line and corresponding border offsets. Using these properties, SEM can reconstruct a more precise polygon expression as shown in the right picture of Fig. \ref{fig:intro} $(b)$. SEM described in Sec.~\ref{sec:sem} can effectively fit text of arbitrary shapes, i.e., those in horizontal, multi-oriented, curved and wavy forms.

The contributions of this work are summarized as follows:
(1) We propose an iterative refinement module which improves the performance of long scene text detection;
(2) An instance-level shape expression module is introduced to solve the problem of detecting scene text of arbitrary shapes;
(3) LOMO with iterative refinement and shape expression modules can be trained in an end-to-end manner and achieves state-of-the-art performance on several benchmarks including text instances of different forms (oriented, long, multi-lingual and curved).

\begin{figure*}
\centering
\includegraphics[scale=0.7]{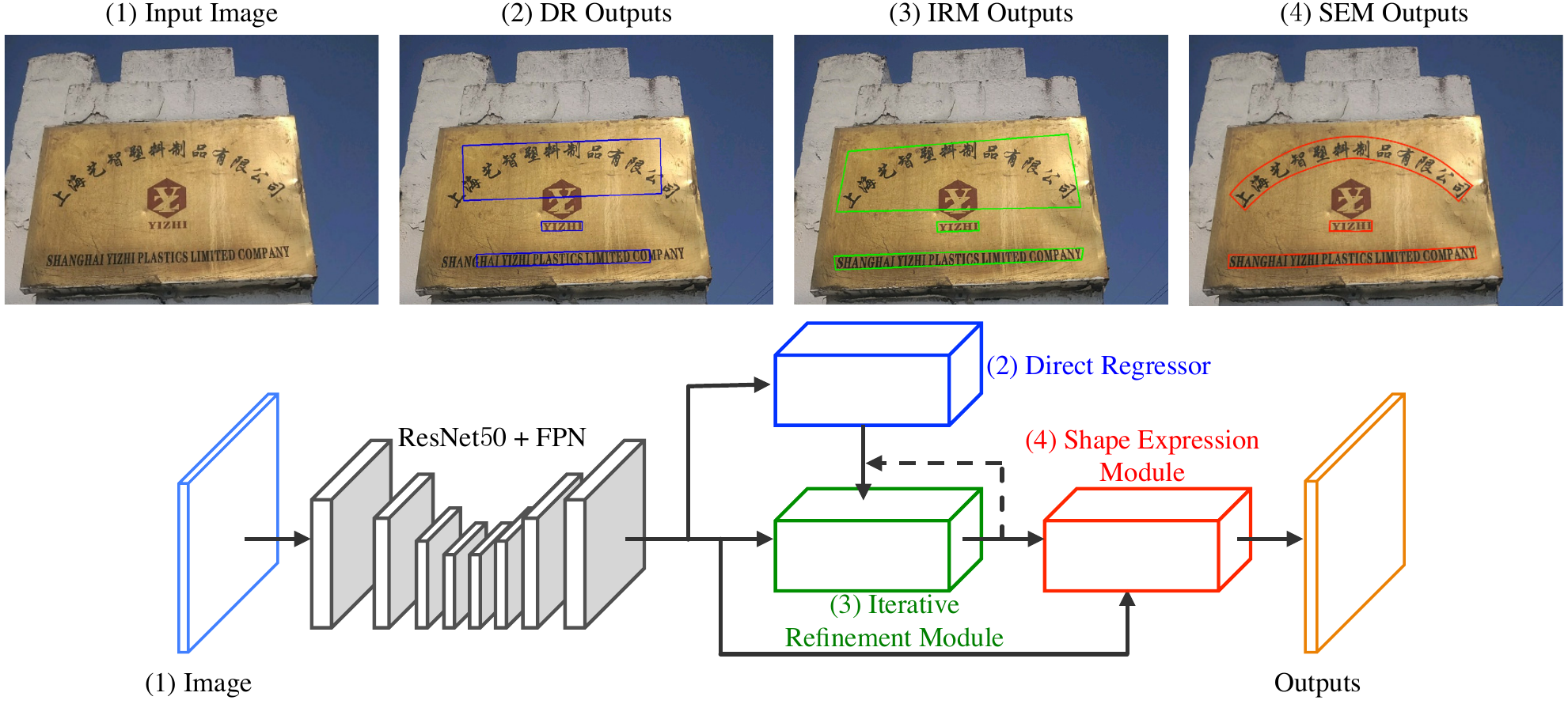}
\caption{The proposed architecture.}
\label{fig:arch}
\vspace{-0.6cm}
\end{figure*}

\section{Related Work}
With the popularity of deep learning, most of the recent scene text detectors are based on deep neural networks. 
According to the basic element of text they handle in natural scene images, these detectors can be roughly classified into three categories: the component-based, detection-based, and segmentation-based approaches.

~\textbf{Component-based methods} \cite{yao2012detecting,tian2016detecting,ren2015faster,shi2017detecting,hu2017wordsup,tian2017wetext,liu2018learning} first detect individual text parts or characters, and then group them into words with a set of post-processing steps. CTPN~\cite{tian2016detecting} adopted the framework of Faster R-CNN~\cite{ren2015faster} to generate dense and compact text components. In ~\cite{shi2017detecting}, scene text is decomposed into two detectable elements namely text segments and links, where a link can indicate whether two adjacent segments belong to the same word and should be connected together. WordSup~\cite{hu2017wordsup} and Wetext~\cite{tian2017wetext} proposed two different weakly supervised learning methods for the character detector, which greatly ease the difficulty of training with insufficient character-level annotations. Liu et al.~\cite{liu2018learning} converted a text image into a stochastic flow graph and then performed Markov Clustering on it to predict instance-level bounding boxes. However, such methods are not robust in scenarios with complex background due to the limitation of staged word/line generation.

~\textbf{Detection-based methods}~\cite{liao2017textboxes,liao2018rotation,ma2018arbitrary,zhou2017east,he2017deep} usually adopt some popular object detection frameworks and models under the supervision of the word or line level annotations. 
TextBoxes~\cite{liao2017textboxes} and RRD~\cite{liao2018rotation} adjusted the anchor ratios of SSD~\cite{Liu2016SSDSS} to handle different aspect ratios of text. 
RRPN~\cite{ma2018arbitrary} proposed rotation region proposal to cover multi-oriented scene text. 
However, EAST~\cite{zhou2017east} and Deep Regression~\cite{he2017deep} directly detected the quadrangles of words in a per-pixel manner without using anchors and proposals. 
Due to their end-to-end design, these approaches can maximize word-level annotation and easily achieve high performance on standard benchmarks. Because the huge variance of text aspect ratios (especially for non-Latin text), as well as the limited receptive filed of the CNN, these methods cannot efficiently handle long text.

~\textbf{Segmentation-based methods}~\cite{zhang2016multi,wu2017self,long2018textsnake,li2018shape} mainly draw inspiration from semantic segmentation methods and regard all the pixels within text bounding boxes as positive regions. The greatest benefit of these methods is the ability to extract arbitrary-shape text. Zhang et al.~\cite{zhang2016multi} first used FCN~\cite{long2015fully} to extract text blocks and then hunted text lines with the statistical information of MSERs~\cite{neumann2010method}. To better separate adjacent text instances, ~\cite{wu2017self} classified each pixel into three categories: non-text, text border and text. TextSnake~\cite{long2018textsnake} and PSENet~\cite{li2018shape} further provided a novel heat map, namely, text center line map to separate different text instances. These methods are based on proposal-free instance segmentation whose performances are strongly affected by the robustness of segmentation results. 


Our method integrates the advantages of detection-based and segmentation-based methods. We propose LOMO which mainly consists of an Iterative Refinement Module (IRM) and a Shape Expression Module (SEM). 
IRM can be inserted into any one-shot text detector to deal with the difficulty of long text detection.
Inspired by Mask R-CNN \cite{he2017mask}, we introduce SEM to handle the arbitrary-shape text. SEM is a region-based method which is more efficient and robust than the region-based methods mentioned above.

\section{Approach}
In this section, we describe the framework of LOMO in detail. First, we briefly introduce the pipeline of our approach to give a tangible concept about \emph{look more than once}. Next, we elaborate all the core modules of LOMO including a direct regressor (DR), iterative refinement module (IRM) and shape expression module (SEM). Finally, the details of training and inference are presented.

\subsection{Overview}
The network architecture of our approach is illustrated in Fig.~\ref{fig:arch}. 
The architecture can be divided into four parts. First, we extract the shared feature maps for three branches including DR, IRM and SEM by feeding the input image to a backbone network.
Our backbone network is ResNet50~\cite{he2016deep} with FPN~\cite{lin2017feature}, where the feature maps of stage-2, stage-3, stage-4 and stage-5 in ResNet50 are effectively merged. Therefore, the size of shared feature maps is $1/4$ of the input image, and the channel number is 128.
Then, we adopt a direct regression network that is similar to EAST~\cite{zhou2017east} and Deep Regression~\cite{he2017deep} as our direct regressor (DR) branch to predict word or text-line quadrangle in a per-pixel manner. 
Usually, the DR branch falls quite short of detecting extremely long text as shown by blue quadrangles in Fig.~\ref{fig:arch} (2), due to the limitation of receptive field. 
Therefore, the next branch namely IRM is introduced to settle this problem. IRM can iteratively refine the input proposals from the outputs of DR or itself to make them closer to the ground-truth bounding box. The IRM described in Sec.~\ref{sec:irm} can perform the refinement operation once or several times, according to the needs of different scenarios. 
With the help of IRM, the preliminary text proposals are refined to cover text instances more completely, as green quadrangles shown in Fig.~\ref{fig:arch} (3).
Finally, in order to obtain tight representation, especially for irregular text, in which the proposal form of  quadrangle easily cover much background region, the SEM reconstruct the shape expression of text instance by learning its geometry attributes including text region, text center line and border offsets (distance between center line and upper/lower border lines). The details of SEM are presented at Sec.~\ref{sec:sem}, and the red polygons shown in Fig.~\ref{fig:arch} (4) are the intuitive visual results.

\subsection{Direct Regressor}
Inspired by \cite{zhou2017east}, a fully convolutional sub-network is adopted as the text direct regressor. 
Based on the shared feature maps,  a dense prediction channel of text/non-text is calculated to indicate the pixel-wise confidence of being text. Similar to \cite{zhou2017east}, pixels in shrunk version of the original text regions are considered positive. 
For each positive sample, $8$ channels predict its offset values to $4$ corner of the quadrangle containing this pixel. The loss function of  DR branch consists of two terms: text/non-text classification term and location regression term.

We regard the text/non-text classification term as a binary segmentation task on the 1/4 down-sampled score map. Instead of using dice-coefficient loss~\cite{milletari2016v} directly, we propose a scale-invariant version for improving the scale generalization of DR in detecting text instances under the receptive field size. The scale-invariant dice-coefficient function is defined as:
\begin{equation}
L_{cls}=1-\frac{2 * sum (y \cdot \hat{y} \cdot w)}{sum(y \cdot w) + sum(\hat{y} \cdot w)},
\label{equa:scale-dice}
\end{equation}
where $y$ is a 0/1 label map, $\hat{y}$ is the predicted score map, and $sum$ is a cumulative function on 2D space. Besides, $w$ in Eq.~\ref{equa:scale-dice} is a 2D weight map. The values of positive positions are calculated by a normalized constant $l$ dividing the shorter sides of the quadrangles they belong to, while the values of negative positions are set to $1.0$. We set $l$ to $64$ in our experiments.

Besides, we adopt the smooth $L_1$ loss~\cite{ren2015faster} to optimize the location regression term $L_{loc}$.
Combining these two terms together, the overall loss function of DR can be written as:
\begin{equation}
L_{dr}= \lambda L_{cls}+L_{loc},
\end{equation}
where a hyper-parameter $\lambda $ balances two loss terms, which is set to $0.01$ in our experiments.

\subsection{Iterative Refinement Module}\label{sec:irm}
\begin{figure}
\centering
\includegraphics[scale=0.45]{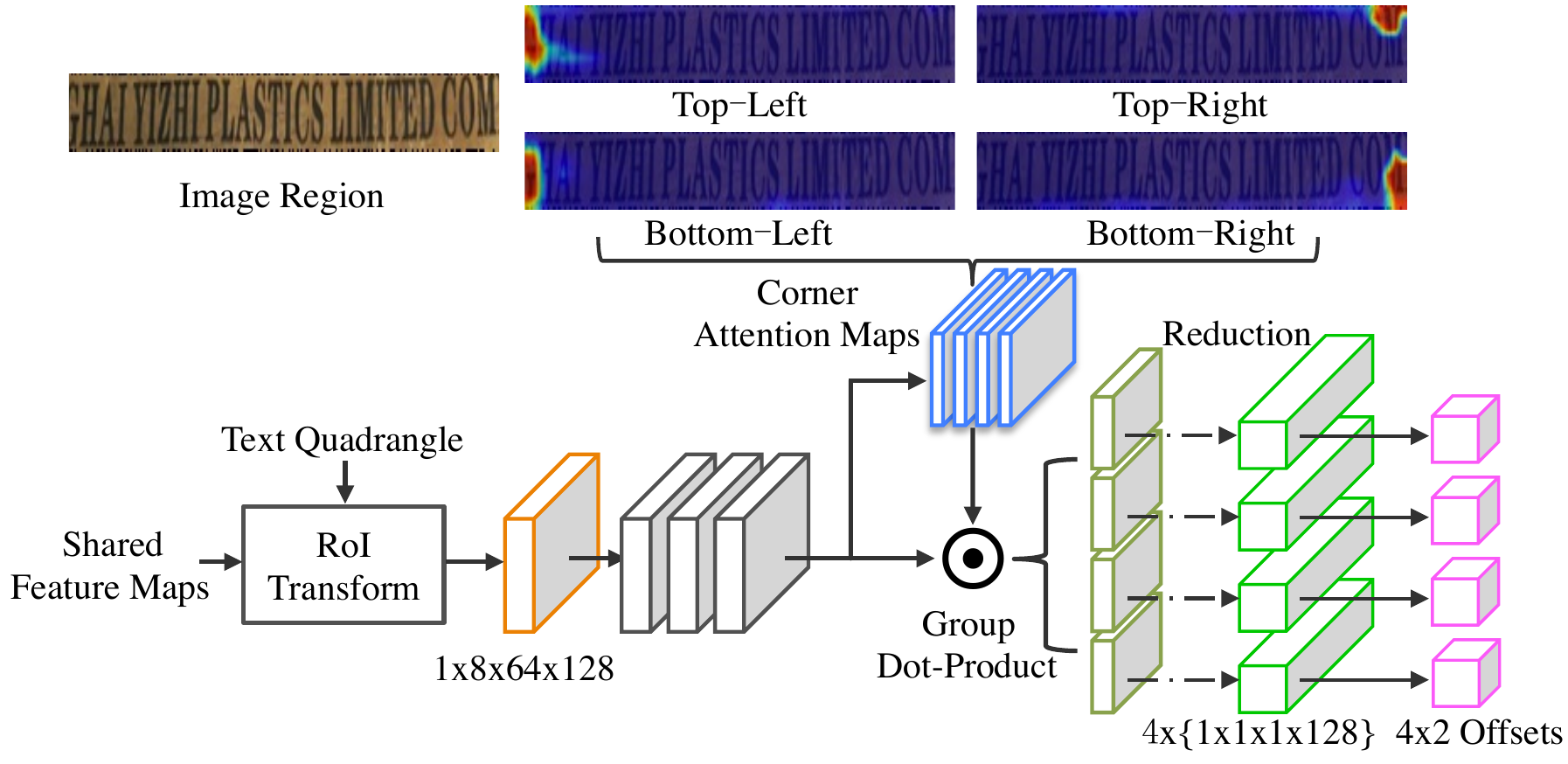}
\caption{The visualization of IRM.}
\label{fig:irm}
\vspace{-0.5cm}
\end{figure}
The desgin of IRM inherits from the region-based object detector~\cite{ren2015faster} with only the boundbing box regression task. However, we use RoI transform layer~\cite{sun2018textnet} to extract the feature block of the input text quadrangle instead of RoI pooling~\cite{ren2015faster} layer or RoI align~\cite{he2017mask} layer. Compared to the latter two ones, the former one can extract the feature block of quadrangle proposal while keeping the aspect ratio unchanged. Besides, as analyzed in Sec.~\ref{sec:intro} that the location close to the corner points can perceive more accurate boundary information within the same receptive field. Thus, a corner attention mechanism is introduced to regress the coordinate offsets of each corner.

The detailed structure is shown in Fig.~\ref{fig:irm}. For one text quadrangle, we feed it with the shared feature maps to RoI transform layer, and then a $1\times8\times64\times128$ feature block is obtained. Afterwards, three $3\times3$ convolutional layers are followed to further extract rich context, namely $f_r$. Next, we use a $1\times1$ convolutional layer and a sigmoid layer to  automatically learn $4$ corner attention maps named $m_a$. The values at each corner attention map denote the contribution weights to support the offset regression of the corresponding corner. With $f_r$ and $m_a$, $4$ corner regression features can be extracted by group dot production and sum reduction operation:
\begin{equation}
f^i_c = reduce\_sum(f_r \cdot m^i_a, axis=[1,2])|i=1,...,4
\end{equation}
where $f^i_c$ denotes the $i$-th corner regression feature whose shape is $1\times1\times1\times128$, $m^i_a$ is the $i$-th learned corner attention map.
Finally, $4$ headers (each header consists of two $1\times1$ convolutional layers) are applied to predict the offsets of $4$ corners between the input quadrangle and the ground-truth text box based on the corner regression features $f_c$.

In training phase, we keep $K$ preliminary detected quadrangles from DR and the corner regression loss can be represented by:
\begin{equation}
L_{irm}= \frac{1}{K*8}\sum_{k=1}^{K}\sum_{j=1}^{8}smooth_{L_{1}}\left ( 
c^{j}_k - \hat{c^{j}_k} \right ),
\end{equation}
where $c^{j}_k$ means the $j$-th coordinate offset between the $k$-th pair of detected quadrangle and ground-truth quadrangle, and $\hat{c^{j}_k}$ is the corresponding predicted value.
As shown in Fig. \ref{fig:irm}, the strong response on four corner attention maps represent the high support for the respective corner regression. By the way, IRM can perform refinement once or more times during testing, if it can bring benefits successively.

\subsection{Shape Expression Module}\label{sec:sem}
\begin{figure*}
\centering
\includegraphics[scale=0.7]{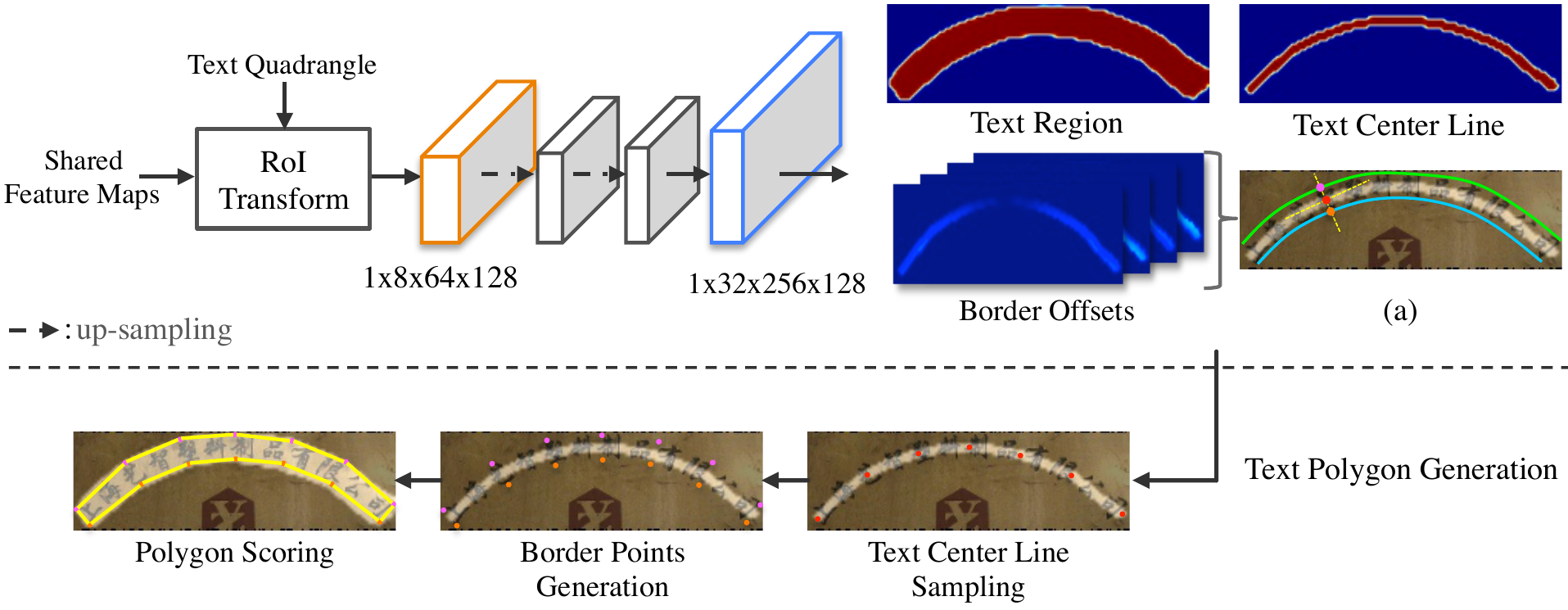}
\caption{The visualization of SEM.}
\label{fig:sem}
\vspace{-0.6cm}
\end{figure*}
The text expression of quadrangle fails to precisely describe the text instances of irregular shapes, especially curved or wavy shapes as shown in Fig.~\ref{fig:intro} (b). Inspired by Mask R-CNN \cite{he2017mask}, we propose a proposal-based shape expression module (SEM) to solve this problem. 
SEM is a fully convolutional network followed with a RoI transform layer.
Three types of text geometry properties including text region, text center line and border offsets (offsets between text center line and upper/lower text border lines) are regressed in SEM to reconstruct the precise shape expression of text instance.
\textbf{Text region} is a binary mask, in which foreground pixels (i.e., those within the polygon annotation) are marked as $1$ and background pixels $0$. \textbf{Text center line} is also a binary mask based on the side-shrunk version of text polygon annotation. 
\textbf{Border offsets} are $4$ channel maps, which have valid values within the area of positive response on the corresponding location of the text line map.
As the center line sample (red point) shows in Fig.~\ref{fig:sem} (a), we draw a normal line that is perpendicular to its tangent, and this normal line is intersected with the upper and lower border line to get two border points (i.e., pink and orange ones). 
For each red point, the 4 border offsets are obtained by calculating the distance from itself to its two related border points.

The structure of SEM is illustrated in Fig.~\ref{fig:sem}, two convolutional stages (each stage consists of one up-sampling layer and two $3\times3$ convolution layers) are followed with the feature block extracted by RoI transform layer, then we use one $1\times1$ convolutional layer with 6 output channels to regress all the text property maps. The objective function of SEM is defined as follows:
\begin{equation}
L_{sem}= \frac{1}{K} \sum^{K} \left( \lambda _{1}L_{tr}+\lambda _{2}L_{tcl}+\lambda _{3}L_{border} \right),
\label{equa:sem}
\end{equation}
where $K$ denotes the number of text quadrangles kept from IRM, $L_{tr}$ and $L_{tcl}$ are dice-coefficient loss for text region and text center line respectively, $L_{border}$ are calculated by smooth $L_{1}$ loss~\cite{ren2015faster}. The weights $\lambda_1$, $\lambda_2$ and $\lambda_3$ are set to $0.01$, $0.01$ and $1.0$ in our experiments.

\textbf{Text Polygon Generation:} We propose a flexible text polygon generation strategy to reconstruct text instance expression of arbitrary shapes, as shown in Fig.~\ref{fig:sem}. The strategy consists of three steps: text center line sampling, border points generation and polygon scoring. 
Firstly, in the process of center line sampling, 
we sample $n$ points at equidistance intervals from left to right on the predicted text center line map. 
According to the label definition in the SCUT-CTW1500~\cite{yuliang2017detecting}, 
we set $n$ to 7 in the curved text detection experiments~\ref{exp:ctw} and to 2 when dealing with text detection in such benchmarks~\cite{karatzas2015icdar,nayef2017icdar2017,shi2017icdar2017} labeled with quadrangle annotations considering the dataset complexity. 
Afterwards, we can determine the corresponding border points based on the sampled center line points, considering the information provided by $4$ border offset maps in the same location. As illustrated in Fig.~\ref{fig:sem} (Border Points Generation), $7$ upper border points (pink) and $7$ lower border points (orange) are obtained. 
By linking all the border points clockwise, we can obtain a complete text polygon representation. Finally, we compute the mean value of the text region response within the polygon as new confidence score.

\subsection{Training and Inference}\label{sec:train_infer}
We train the proposed network in an end-to-end manner using the following loss function:
\begin{equation}
L= \gamma _{1}L_{dr}+\gamma _{2}L_{irm}+\gamma _{3}L_{sem}, \label{con:allloss}
\end{equation}
where $L_{dr}$, $L_{irm}$ and $L_{sem}$ represent the loss of DR, IRM and SEM, respectively. The weights $\gamma _{1}$, $\gamma _{2}$, and $\gamma _{3}$ trade off among three modules and are all set to $1.0$ in our experiments.

Training is divided into two stages: warming-up and fine-tuning. At warming-up step, we train DR branch using synthetic dataset~\cite{gupta2016synthetic} only for 10 epochs. 
In this way, DR can generate high-recall proposals to cover most of text instance in real data.
At fine-tuning step, we fine-tune all three branches on real datasets including ICDAR2015~\cite{karatzas2015icdar}, ICDAR2017-RCTW~\cite{shi2017icdar2017}, SCUT-CTW1500~\cite{yuliang2017detecting}, Total-Text~\cite{ch2017total} and ICDAR2017-MLT~\cite{nayef2017icdar2017} about another $10$ epochs. Both IRM and SEM branches use the same proposals which generated by DR branch. Non-Maximum Suppression (NMS) is used to keep top $K$ proposals. Since the DR performs poorly at first, which will affect the convergence of IRM and SEM branches, we replace $50\%$ of the top $K$ proposals with the randomly disturbed GT text quadrangles in practice.
Note that IRM performs refinement only once during training.

In inference phase, DR firstly generates score map and geometry maps of quadrangle, and NMS is followed to generate preliminary proposals. 
Next, both the proposals and the shared feature maps are feed into IRM for multiple refinements.
The refined quadrangles and shared feature maps are feed into SEM to generate the precise text polygons and confidence scores. Finally, a threshold $s$ is used to remove low-confidence polygons. We set $s$ to 0.1 in our experiments.

\section{Experiments} 

To compare LOMO with existing state-of-the-art methods, we perform thorough experiments on five public scene text detection datasets, i.e., ICDAR 2015, ICDAR2017-MLT,  ICDAR2017-RCTW, SCUT-CTW1500 and Total-Text. The evaluation protocols are based on \cite{karatzas2015icdar,nayef2017icdar2017,shi2017icdar2017,yuliang2017detecting,ch2017total} respectively.
\subsection{Datasets}

The datasets used for the experiments in this paper are briefly introduced below:

\textbf{ICDAR 2015}. The ICDAR 2015 dataset~\cite{karatzas2015icdar} is collected for the ICDAR 2015 Robust Reading Competition, with $1000$ natural images for training and $500$ for testing. The images are acquired using Google Glass and the text accidentally appears in the scene. The ground truth is annotated with word-level quadrangle.

\textbf{ICDAR2017-MLT}. The ICDAR2017-MLT~\cite{nayef2017icdar2017} is a large scale multi-lingual text dataset, which includes $7200$ training images, $1800$ validation images and $9000$ test images. The dataset consists of scene text images which come from $9$ languages. 
The text regions in ICDAR2017-MLT are also annotated by $4$ vertices of the quadrangle.

\textbf{ICDAR2017-RCTW}. The ICDAR2017-RCTW~\cite{shi2017icdar2017} comprises $8034$ training images and $4229$ test images with scene texts printed in either Chinese or English. The images are captured from different sources including street views, posters, screen-shot, etc. Multi-oriented words and text lines are annotated using quadrangles.
  
\textbf{SCUT-CTW1500}. The SCUT-CTW1500~\cite{yuliang2017detecting} is a challenging dataset for curved text detection. It consists of $1000$ training images and $500$ test images. Different from traditional dataset (e.g., ICDAR 2015, ICDAR2017-MLT), the text instances in SCUT-CTW1500 are labelled by polygons with $14$ vertices.

\textbf{Total-Text}. The Total-Text~\cite{ch2017total} is another curved text benchmarks, which consist of $1255$ training images and $300$ testing images. Different from SCUT-CTW1500, the annotations are labelled in word-level.

\begin{figure*}
\centering
\includegraphics[scale=0.73]{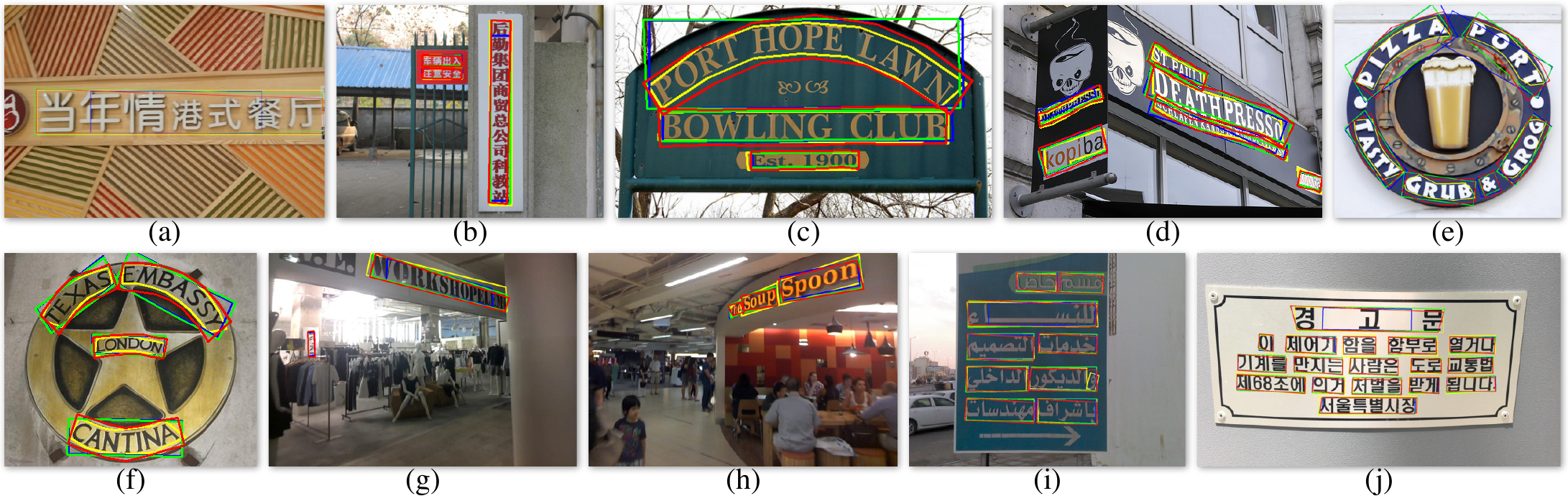}
\caption{The visualization of detection results. (a)~(b) are sampled from ICDAR2017-RCTW, (c)~(d) are from SCUT-CTW1500, (e)~(f) are from Total-Text, (g)~(h) are from ICDAR2015, and (i)~(j) are from ICDAR2017-MLT. The yellow polygons are ground truth annotations. The localization quadrangles in blue and in green represent the detection results of DR and IRM respectively. The  contours in red are the detection results of SEM.}
\label{fig:vis_res}
\vspace{-0.3cm}
\end{figure*}
\subsection{Implementation Details}
The training process is divided into two steps as described in Sec.~\ref{sec:train_infer}. In the warming-up step, we apply adam optimizer to train our model with learning rate $10^{-4}$, and the learning rate decay factor is 0.94. In the fine-tuning step, the learning rate is re-initiated to $10^{-4}$. For all datasets, we randomly crop the text regions and resize them to $512\times512$. The cropped image regions will be rotated randomly in 4 directions including $0^{\circ}$, $90^{\circ}$, $180^{\circ}$ and $270^{\circ}$. All the experiments are performed on a standard workstation with the following configuration, CPU: Intel(R) Xeon(R) CPU E5-2620 v2 @ 2.10GHz x16; GPU: Tesla K40m; RAM: 160 GB. During the training time, we set the batch size to $8$ on $4$ GPUs in parallel and the number $K$ of detected proposals generated by DR branch is set to $24$ per gpu. In inference phase, the batch size is set to $1$ on $1$ GPU. The full time cost of predicting an image whorse longer size is resized to 512 with keeping original aspect ratio is 224 millisecond.

\subsection{Ablation Study}

\begin{figure}
\centering
\includegraphics[scale=0.45]{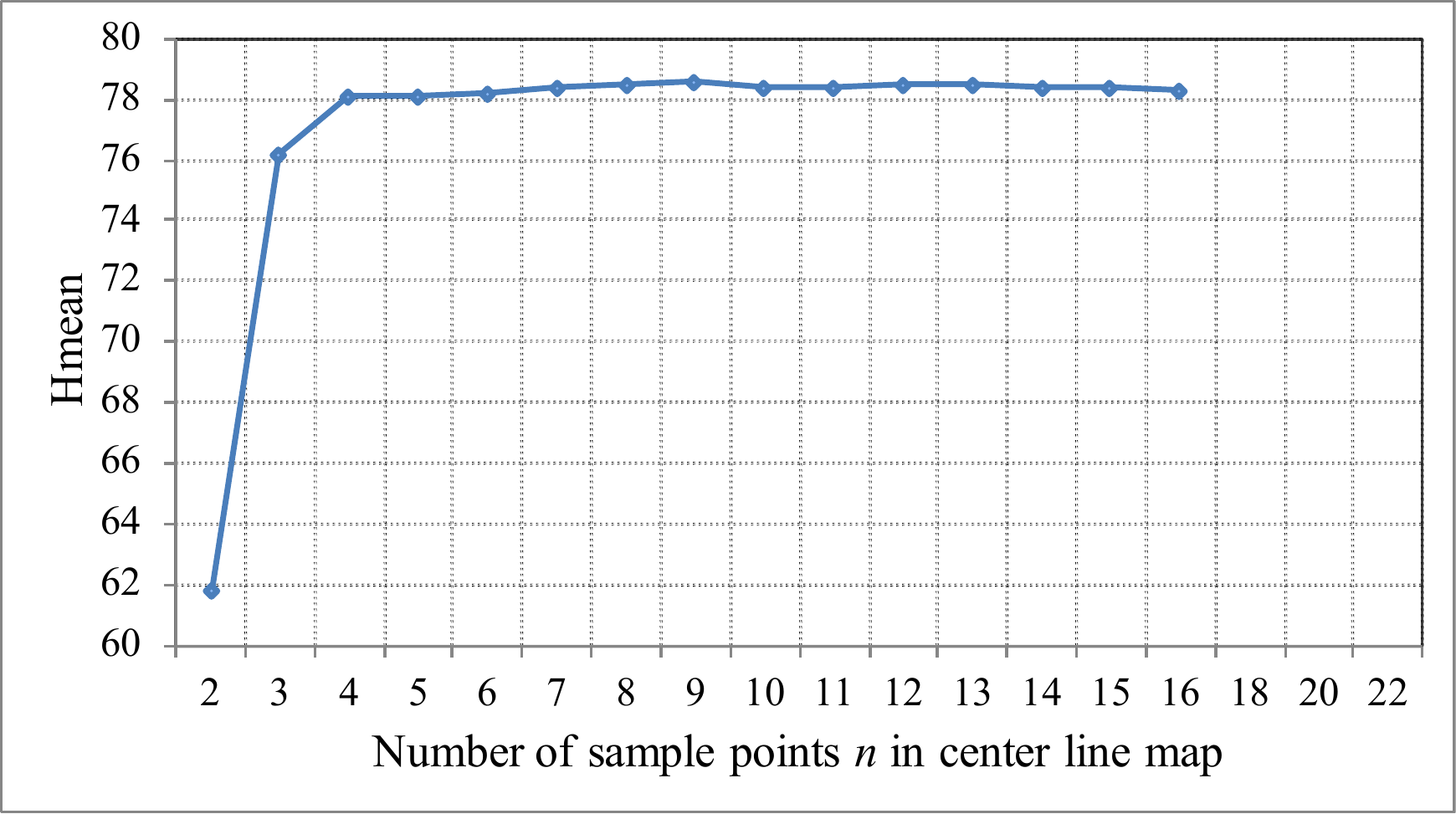}
\caption{Ablations for the number of sample points in center line.}
\label{fig:num_sample_pts}
\vspace{-0.3cm}
\end{figure}

We conduct several ablation experiments to analyze LOMO. The results are shown in Tab.~\ref{tab:irm}, Tab.~\ref{tab:att_map}, Fig.~\ref{fig:num_sample_pts} and Tab.~\ref{tab:sem}. The details are discussed as follows.


\textbf{Discussions about IRM:} An evaluation of IRM on ICDAR2017-RCTW~\cite{shi2017icdar2017} is shown in Tab.~\ref{tab:irm}. We use Resnet50-FPN as backbone, and fix the longer side of text images to 1024 while keeping the aspect ratio. As can be seen in Tab.~\ref{tab:irm}, IRM achieves successive gains of $3.34\%$, $3.66\%$ and $3.77\%$ in Hmean with RT set to $1$, $2$ and $3$ respectively, compared to DR branch without IRM. This shows the great effectiveness of IRM for refining long text detection. It is possible to obtain better performance with even more refinement times in this way. To preserve fast inference time, we set RT to $2$ in the remaining experiments.

\textbf{Corner Attention Map:} LOMO utilizes corner attention map in IRM. For fair comparison, we generate a model based on DR+IRM without corner attention map and evaluate its performance on ICDAR2017-RCTW~\cite{shi2017icdar2017}. The results are shown in Tab.~\ref{tab:att_map}. We can see that DR+IRM without corner attention map leads to a loss of $0.3\%$ and $1.5\%$ in Hmean in the protocol of IoU$@0.5$ and IoU$@0.7$ respectively. This suggests that the corner features enhanced by corner attention map help to detect long text instances.


\textbf{Benefits from SEM:} We evaluate the benefits of SEM on SCUT-CTW1500~\cite{yuliang2017detecting} in Tab.~\ref{tab:sem}. Methods (a) and (b) are based on DR branch without IRM. We resize the longer side of text images to 512 and keep the aspect ratio unchanged. As demonstrated in Tab.~\ref{tab:sem}, SEM significantly improves the Hmean by $7.17\%$. We also conduct experiments of SEM based on DR with IRM in methods (c)and (d). Tab.~\ref{tab:sem} shows that SEM improves the Hmean by a large margin ($6.34\%$). The long-standing challenge of curved text detection is largely resolved by SEM.

\textbf{Number of Sample Points in Center Line:} LOMO performs text polygon generation step to output final detection results, which are decided by the number of sample points $n$ in center lines. We evaluate the performance of LOMO with several different $n$ on SCUT-CTW1500~\cite{yuliang2017detecting}. As illustrated in Fig.~\ref{fig:num_sample_pts}, the Hmean of LOMO increases significantly from $62\%$ to $78\%$ and then converges when $n$ is chosen from $2$ to $16$. 
We set $n$ to $7$ in our remaining experiments.

\begin{table}[]
\small
\centering
\caption{Ablations for refinement times (RT) of IRM.} \label{tab:irm}
\begin{tabular}{c|c|ccc|c}
\hline
Method & RT & Recall & Precision & Hmean & FPS \\ \hline
DR            & 0 & 49.09   & 73.80      & 58.96  & 4.5    \\
DR+IRM & 1 & 51.25   & 79.42      & 62.30  & 3.8    \\
DR+IRM & 2 & 51.42   & 80.07      & 62.62  & 3.4    \\
DR+IRM & 3 & 51.48   & 80.29      & 62.73  & 3.0    \\ \hline
\end{tabular}
\vspace{-0.1cm}
\end{table}

\begin{table}[]
\small
\centering
\caption{Ablations for Corner Attention Map (CAM). The study is based on DR+IRM with RT set to $2$.} \label{tab:att_map}
\setlength{\tabcolsep}{1mm}{
\begin{tabular}{c|ccc|ccc}
\hline
Protocol & \multicolumn{3}{c|}{IoU$@0.5$} & \multicolumn{3}{c}{IoU$@0.7$} \\ \hline
Method   & R & P & H & R & P & H \\ \hline
w.o. CAM  & 51.09          & 79.85          & 62.31          & 42.34          & 66.17          & 51.64 \\
with CAM  & \textbf{51.42} & \textbf{80.07} & \textbf{62.62} & \textbf{43.64} & \textbf{67.95} & \textbf{53.14} \\ \hline 
\end{tabular}}
\vspace{-0.2cm}
\end{table}

\begin{table}[]
\small
\centering
\caption{Ablation study for SEM} \label{tab:sem}
\setlength{\tabcolsep}{1mm}{
\begin{tabular}{c|cc|ccc|c}
\hline
Method & IRM & SEM & Recall & Precision & Hmean & FPS\\ \hline
(a)  &            &            & 63.10       & 80.07          & 70.58      & 11.9    \\
(b)  &            & \checkmark & 69.20       & 88.72          & 77.75      & 6.4     \\ \hline
(c)  & \checkmark &            & 64.24       & 82.13          & 72.09      & 6.3     \\
(d)  & \checkmark & \checkmark & 69.62       & 89.79          & 78.43      & 4.4     \\ \hline
\end{tabular}}
\vspace{-0.3cm}
\end{table}

\subsection{Evaluation on Long Text Benchmark}
We evaluate the performance of LOMO for detecting long text cases on ICDAR2017-RCTW. During training, we use all of the training images of this dataset in fine-tuning step. For single-scale testing, if the longer side of an input image is larger than 1024, we resize the longer side of the image to $1024$ and keep the aspect ratio. For multi-scale testing, the longer side scales of resized images include $512$, $768$, $1024$, $1536$ and $2048$. The quantitative results are listed in Tab. \ref{tab:ic17rctw}. LOMO achieves $62.3\%$ in Hmean, surpassing the best single scale method RRD by $6.6\%$. Thanks to multi-scale testing, LOMO MS further improves the Hmean to $68.4\%$, which is state-of-the-art on this benchmark. 
Some detection results of LOMO are shown in Fig. \ref{fig:vis_res} (a) and (b). LOMO achieves promising results in detecting extremely long text. In Fig. \ref{fig:vis_res} (a) and (b), we compare the detection results of DR and IRM. DR shows limited ability to detect long text, while the localization quadrangles of IRM can perceive more complete text regions.



\begin{table}[]
\centering
\caption{Quantitative results of different methods on ICDAR2017-RCTW. MS denotes multi-scale testing.} \label{tab:ic17rctw}
\setlength{\tabcolsep}{1mm}{
\begin{tabular}{c|ccc}
\hline
Method       & Recall & Precision & Hmean \\ \hline 
Official baseline \cite{shi2017icdar2017} & 40.4          & 76.0           & 52.8         \\ 
EAST \cite{zhou2017east}                  & 47.8          & 59.7           & 53.1         \\ 
RRD \cite{liao2018rotation}               & 45.3          & 72.4           & 55.7         \\ 
RRD MS \cite{liao2018rotation}            & 59.1          & 77.5           & 67.0         \\ 
Border MS \cite{xue2018accurate}          & 58.8          & 78.2           & 67.1         \\ \hline 
\textbf{LOMO}                             & 50.8          & \textbf{80.4} & 62.3          \\ 
\textbf{LOMO MS}                          & \textbf{60.2} & 79.1          & \textbf{68.4} \\ \hline 
\end{tabular}}
\vspace{-0.4cm}
\end{table}

\subsection{Evaluation on Curved Text Benchmark} \label{exp:ctw} 
We evaluate the performance of LOMO on SCUT-CTW1500 and Total-Text, which contains many arbitrarily shaped text instances, to validate the ability of detecting arbitrary-shape text. During training, we stop the fine-tuning step at about $10$ epochs, using training images only. For testing, the number of sample points in center line is set to $7$, so we can generate text polygons with $14$ vertices. 
All the quantitative results are shown in Tab.~\ref{tab:ctw}. 
With the help of SEM, LOMO achieves the state-of-the-art results both on SCUT-CTW1500 and Total-Text, and outperform the existing methods (e.g., CTD+TLOC~\cite{yuliang2017detecting}, TextSnake~\cite{long2018textsnake}) by a considerable margin.
In addition, multi-scale testing can further improves Hmean by $2.4\%$ and $1.7\%$ on SCUT-CTW1500 and Total-Text respectively. 
The visualization of curved text detection are shown in Fig. \ref{fig:vis_res} (c) (d) (e) (f). LOMO shows great robustness in detecting arbitrarily curved text instances. It should be noted that the polygons generated by SEM can cover the curved text instances more precisely, compared with the quadrangles of DR and IRM.
\begin{table}[]
\small
\centering
\caption{Quantitative results of different methods on SCUT-CTW1500 and Total-Text. ``R'', ``P'' and ``H'' represent recall, precision and Hmean respectively. Note that EAST is not fine-tuned in these two datasets and the results of it are just for reference.} \label{tab:ctw}
\setlength{\tabcolsep}{1mm}{
\begin{tabular}{c|ccc|ccc}
\hline
Datasets     & \multicolumn{3}{c|}{SCUT-CTW1500}  & \multicolumn{3}{c}{Total-Text}   \\ \hline
Method       & R & P & H  & R & P & H  \\ \hline
DeconvNet \cite{ch2017total}         & - & - & -                     & 40.0          & 33.0          & 36.0 \\
CTPN \cite{tian2016detecting}        & 53.8          & 60.4          & 56.9          & - & - & -     \\
EAST    \cite{zhou2017east}          & 49.1          & 78.7          & 60.4          & 36.2          & 50.0          & 42.0 \\
Mask TextSpotter \cite{lyu2018mask}  & - & - & -                     & 55.0          & 69.0          & 61.3 \\ 
CTD \cite{yuliang2017detecting}      & 65.2          & 74.3          & 69.5          & - & - & -    \\
CTD+TLOC \cite{yuliang2017detecting} & 69.8          & 74.3          & 73.4          & - & - & -    \\
SLPR \cite{zhu2018sliding}           & 70.1          & 80.1          & 74.8          & - & - & -    \\
TextSnake \cite{long2018textsnake}   & \textbf{85.3} & 67.9          & 75.6          & 74.5          & 82.7          & 78.4 \\ \hline
\textbf{LOMO}                        & 69.6          & \textbf{89.2} & 78.4          & 75.7          & \textbf{88.6} & 81.6 \\
\textbf{LOMO MS}                     & 76.5          & 85.7          & \textbf{80.8} & \textbf{79.3} & 87.6          & \textbf{83.3} \\ \hline
\end{tabular}}
\vspace{-0.1cm}
\end{table}

\subsection{Evaluation on Oriented Text Benchmark}
We compare LOMO with the state-of-the-art results on ICDAR 2015 dataset, a standard oriented text dataset. We set the scale of the longer side to $1536$ for single-scale testing. And the longer sides in multi-scale testing are set to $1024$, $1536$ and $2048$. All the results are listed in the Tab.~\ref{tab:ic15}, LOMO outperforms the previous text detection methods which are trained without the help of recognition task, while is on par with end-to-end methods~~\cite{lyu2018mask,buvsta2017deep,sun2018textnet,liu2018fots}. For single-scale testing, LOMO achieves $87.2\%$ Hmean, surpassing all competitors which only use detection training data. Moreover, multi-scale testing increases about $0.5\%$ Hmean. 
Some detection results are shown in Fig. \ref{fig:vis_res} (g) (h). As can be seen, only when detecting long text can IRM achieve significant improvement. It is worthy noting that the detection performance would be further improved if LOMO was equipped with recognition branch in the future.

\begin{table}[]
\centering
\caption{Quantitative results of different methods on ICDAR 2015.} \label{tab:ic15}
\setlength{\tabcolsep}{1mm}{
\begin{tabular}{c|ccc}
\hline
Method       & Recall & Precision & Hmean \\ \hline
SegLink \cite{shi2017detecting}              & 76.5           & 74.7          & 75.6           \\
MCN \cite{liu2018learning}                   & 80.0           & 72.0          & 76.0           \\
SSTD \cite{he2017single}                     & 73.9          & 80.2         & 76.9           \\
WordSup \cite{hu2017wordsup}                 & 77.0           & 79.3          & 78.2           \\
EAST \cite{zhou2017east}                     & 78.3           & 83.3          & 80.7           \\
He et al. \cite{he2017deep}                  & 80.0           & 82.0          & 81.0          \\
TextSnake \cite{long2018textsnake}           & 80.4           & 84.9          & 82.6          \\
PixelLink \cite{deng2018pixellink}           & 82.0           & 85.5          & 83.7          \\
RRD \cite{liao2018rotation}                  & 80.0           & 88.0          & 83.8          \\
Lyu et al. \cite{lyu2018multi}               & 79.7           & 89.5          & 84.3          \\
IncepText \cite{yang2018inceptext}           & 84.3           & 89.4          & 86.8          \\ \hline
Mask TextSpotter \cite{lyu2018mask}          & 81.0           & 91.6          & 86.0          \\
End-to-End TextSpotter \cite{buvsta2017deep} & 86.0           & 87.0          & 87.0          \\
TextNet \cite{sun2018textnet}                & 85.4           & 89.4          & 87.4          \\
FOTS MS \cite{liu2018fots}                   & \textbf{87.9}  & \textbf{91.9} & \textbf{89.8} \\ \hline
\textbf{LOMO}                                & 83.5           & 91.3          & 87.2 \\
\textbf{LOMO MS}                             & 87.6           & 87.8          & 87.7 \\ \hline
\end{tabular}}
\vspace{-0.4cm}
\end{table}

\subsection{Evaluation on Multi-Lingual Text Benchmark}
In order to verify the generalization ability of LOMO on multilingual scene text detection, we evaluate LOMO on ICDAR2017-MLT. The detector is finetuned for $10$ epochs based on the SynthText pre-trained model. In inference phase, we set the longer side to $1536$ for single-scale testing, and the longer side scales for multi-scale testing include $512$, $768$, $1536$ and $2048$. As shown in Table \ref{tab:ic17mlt}, LOMO has the performance leading in single-scale testing compared with most of existing methods~\cite{he2018multi,lyu2018multi,xue2018accurate,zhong2018anchor}, expect for Border~\cite{xue2018accurate} and AFN-RPN~\cite{zhong2018anchor} that do not indicate which testing scales are used. Moreover, LOMO achieves state-of-the-art performance ($73.1\%$) in the multi-scale testing mode. In particular, the proposed method outperforms existing end-to-end approaches (i.e., E2E-MLT~\cite{patel2018e2e} and FOTS~\cite{liu2018fots}) at least $2.3\%$. The visualization of multilingual text detection are shown in Fig. \ref{fig:vis_res} (i) (j), the localization quadrangles of IRM could significantly improve detection compared with DR.

\begin{table}[]
\centering
\caption{Quantitative results of different methods on ICDAR2017-MLT.} \label{tab:ic17mlt}
\setlength{\tabcolsep}{1mm}{
\begin{tabular}{c|ccc}
\hline
Method       & Recall & Precision & Hmean \\ \hline
E2E-MLT \cite{patel2018e2e}          & 53.8          & 64.6          & 58.7   \\
He et al. \cite{he2018multi}         & 57.9          & 76.7          & 66.0   \\
Lyu et al. \cite{lyu2018multi}       & 56.6          & \textbf{83.8} & 66.8   \\
FOTS \cite{liu2018fots}              & 57.5          & 81.0          & 67.3  \\
Border \cite{xue2018accurate}        & 62.1          & 77.7          & 69.0   \\
AF-RPN \cite{zhong2018anchor}        & 66.0          & 75.0          & 70.0 \\
FOTS MS \cite{liu2018fots}         & 62.3          & 81.9          & 70.8  \\
Lyu et al. MS \cite{lyu2018multi}  & \textbf{70.6} & 74.3          & 72.4   \\ \hline
\textbf{LOMO}                        & 60.6          & 78.8          & 68.5   \\
\textbf{LOMO MS}                     & 67.2          & 80.2          & \textbf{73.1} \\ \hline
\end{tabular}}
\vspace{-0.5cm}
\end{table}

\section{Conclusion and Future Work}
In this paper, we propose a novel text detection method (LOMO) to solve the problems of detecting extremely long text and curved text. LOMO consists of three modules including DR, IRM and SEM. DR localizes the preliminary proposals of text. IRM refines the proposals iteratively to solve the issue of detecting long text. SEM proposes a flexible shape expression method for describing the geometry property of scene text with arbitrary shapes. The overall architecture of LOMO can be trained in an end-to-end fashion. The robustness and effectiveness of our approach have been proven on several public benchmarks including long, curved or wavy, oriented and multilingual text cases. In the future, we are interested in developing an end-to-end text reading system for text of arbitrary shapes. 
\paragraph{Acknowledgements}
This work was supported in part by the National Natural Science Foundation of China under Grants 61571382, 81671766, 61571005, 81671674, 61671309 and U1605252, and in part by the Fundamental Research Funds for the Central Universities under Grant 20720160075 and 20720180059.

{\small
\bibliographystyle{ieee}
\bibliography{egbib}

\begin{thebibliography}{10}\itemsep=-1pt

\bibitem{buvsta2017deep}
M.~Bu{\v{s}}ta, L.~Neumann, and J.~Matas.
\newblock Deep textspotter: An end-to-end trainable scene text localization and
  recognition framework.
\newblock In {\em ICCV}, pages 2223--2231. IEEE, 2017.

\bibitem{ch2017total}
C.~K. Ch'ng and C.~S. Chan.
\newblock Total-text: A comprehensive dataset for scene text detection and
  recognition.
\newblock In {\em ICDAR}, volume~1, pages 935--942. IEEE, 2017.

\bibitem{deng2018pixellink}
D.~Deng, H.~Liu, X.~Li, and D.~Cai.
\newblock Pixellink: Detecting scene text via instance segmentation.
\newblock {\em arXiv preprint arXiv:1801.01315}, 2018.

\bibitem{gupta2016synthetic}
A.~Gupta, A.~Vedaldi, and A.~Zisserman.
\newblock Synthetic data for text localisation in natural images.
\newblock In {\em CVPR}, pages 2315--2324, 2016.

\bibitem{he2017mask}
K.~He, G.~Gkioxari, P.~Doll{\'a}r, and R.~Girshick.
\newblock Mask r-cnn.
\newblock In {\em ICCV}, pages 2980--2988. IEEE, 2017.

\bibitem{he2016deep}
K.~He, X.~Zhang, S.~Ren, and J.~Sun.
\newblock Deep residual learning for image recognition.
\newblock In {\em CVPR}, pages 770--778, 2016.

\bibitem{he2017single}
P.~He, W.~Huang, T.~He, Q.~Zhu, Y.~Qiao, and X.~Li.
\newblock Single shot text detector with regional attention.
\newblock In {\em ICCV}, volume~6, 2017.

\bibitem{he2017deep}
W.~He, X.-Y. Zhang, F.~Yin, and C.-L. Liu.
\newblock Deep direct regression for multi-oriented scene text detection.
\newblock {\em arXiv preprint arXiv:1703.08289}, 2017.

\bibitem{he2018multi}
W.~He, X.-Y. Zhang, F.~Yin, and C.-L. Liu.
\newblock Multi-oriented and multi-lingual scene text detection with direct
  regression.
\newblock {\em IEEE Transactions on Image Processing}, 27(11):5406--5419, 2018.

\bibitem{hu2017wordsup}
H.~Hu, C.~Zhang, Y.~Luo, Y.~Wang, J.~Han, and E.~Ding.
\newblock Wordsup: Exploiting word annotations for character based text
  detection.
\newblock In {\em ICCV}, 2017.

\bibitem{huang2016detecting}
W.~Huang, D.~He, X.~Yang, Z.~Zhou, D.~Kifer, and C.~L. Giles.
\newblock Detecting arbitrary oriented text in the wild with a visual attention
  model.
\newblock In {\em ACM MM}, pages 551--555. ACM, 2016.

\bibitem{karatzas2015icdar}
D.~Karatzas, L.~Gomez-Bigorda, A.~Nicolaou, S.~Ghosh, A.~Bagdanov, M.~Iwamura,
  J.~Matas, L.~Neumann, V.~R. Chandrasekhar, S.~Lu, et~al.
\newblock Icdar 2015 competition on robust reading.
\newblock In {\em ICDAR}, pages 1156--1160. IEEE, 2015.

\bibitem{li2018shape}
X.~Li, W.~Wang, W.~Hou, R.-Z. Liu, T.~Lu, and J.~Yang.
\newblock Shape robust text detection with progressive scale expansion network.
\newblock {\em arXiv preprint arXiv:1806.02559}, 2018.

\bibitem{liao2017textboxes}
M.~Liao, B.~Shi, X.~Bai, X.~Wang, and W.~Liu.
\newblock Textboxes: A fast text detector with a single deep neural network.
\newblock In {\em AAAI}, pages 4161--4167, 2017.

\bibitem{liao2018rotation}
M.~Liao, Z.~Zhu, B.~Shi, G.-s. Xia, and X.~Bai.
\newblock Rotation-sensitive regression for oriented scene text detection.
\newblock In {\em CVPR}, pages 5909--5918, 2018.

\bibitem{lin2017feature}
T.-Y. Lin, P.~Doll{\'a}r, R.~B. Girshick, K.~He, B.~Hariharan, and S.~J.
  Belongie.
\newblock Feature pyramid networks for object detection.
\newblock In {\em CVPR}, volume~1, page~4, 2017.

\bibitem{Liu2016SSDSS}
W.~Liu, D.~Anguelov, D.~Erhan, C.~Szegedy, S.~E. Reed, C.-Y. Fu, and A.~C.
  Berg.
\newblock Ssd: Single shot multibox detector.
\newblock In {\em ECCV}, 2016.

\bibitem{liu2018fots}
X.~Liu, D.~Liang, S.~Yan, D.~Chen, Y.~Qiao, and J.~Yan.
\newblock Fots: Fast oriented text spotting with a unified network.
\newblock In {\em CVPR}, pages 5676--5685, 2018.

\bibitem{liu2018learning}
Z.~Liu, G.~Lin, S.~Yang, J.~Feng, W.~Lin, and W.~L. Goh.
\newblock Learning markov clustering networks for scene text detection.
\newblock {\em arXiv preprint arXiv:1805.08365}, 2018.

\bibitem{long2015fully}
J.~Long, E.~Shelhamer, and T.~Darrell.
\newblock Fully convolutional networks for semantic segmentation.
\newblock In {\em CVPR}, pages 3431--3440, 2015.

\bibitem{long2018textsnake}
S.~Long, J.~Ruan, W.~Zhang, X.~He, W.~Wu, and C.~Yao.
\newblock Textsnake: A flexible representation for detecting text of arbitrary
  shapes.
\newblock In {\em ECCV}, pages 19--35. Springer, 2018.

\bibitem{lyu2018mask}
P.~Lyu, M.~Liao, C.~Yao, W.~Wu, and X.~Bai.
\newblock Mask textspotter: An end-to-end trainable neural network for spotting
  text with arbitrary shapes.
\newblock In {\em ECCV}, pages 67--83, 2018.

\bibitem{lyu2018multi}
P.~Lyu, C.~Yao, W.~Wu, S.~Yan, and X.~Bai.
\newblock Multi-oriented scene text detection via corner localization and
  region segmentation.
\newblock In {\em CVPR}, pages 7553--7563, 2018.

\bibitem{ma2018arbitrary}
J.~Ma, W.~Shao, H.~Ye, L.~Wang, H.~Wang, Y.~Zheng, and X.~Xue.
\newblock Arbitrary-oriented scene text detection via rotation proposals.
\newblock {\em IEEE Transactions on Multimedia}, 2018.

\bibitem{milletari2016v}
F.~Milletari, N.~Navab, and S.-A. Ahmadi.
\newblock V-net: Fully convolutional neural networks for volumetric medical
  image segmentation.
\newblock In {\em IC3DV}, pages 565--571. IEEE, 2016.

\bibitem{nayef2017icdar2017}
N.~Nayef, F.~Yin, I.~Bizid, H.~Choi, Y.~Feng, D.~Karatzas, Z.~Luo, U.~Pal,
  C.~Rigaud, J.~Chazalon, et~al.
\newblock Icdar2017 robust reading challenge on multi-lingual scene text
  detection and script identification-rrc-mlt.
\newblock In {\em ICDAR}, volume~1, pages 1454--1459. IEEE, 2017.

\bibitem{neumann2010method}
L.~Neumann and J.~Matas.
\newblock A method for text localization and recognition in real-world images.
\newblock In {\em ACCV}, pages 770--783, 2010.

\bibitem{patel2018e2e}
Y.~Patel, M.~Bu{\v{s}}ta, and J.~Matas.
\newblock E2e-mlt-an unconstrained end-to-end method for multi-language scene
  text.
\newblock {\em arXiv preprint arXiv:1801.09919}, 2018.

\bibitem{ren2015faster}
S.~Ren, K.~He, R.~Girshick, and J.~Sun.
\newblock Faster r-cnn: Towards real-time object detection with region proposal
  networks.
\newblock In {\em NeurIPS}, pages 91--99, 2015.

\bibitem{shi2017detecting}
B.~Shi, X.~Bai, and S.~Belongie.
\newblock Detecting oriented text in natural images by linking segments.
\newblock In {\em CVPR}, pages 3482--3490. IEEE, 2017.

\bibitem{shi2017icdar2017}
B.~Shi, C.~Yao, M.~Liao, M.~Yang, P.~Xu, L.~Cui, S.~Belongie, S.~Lu, and
  X.~Bai.
\newblock Icdar2017 competition on reading chinese text in the wild (rctw-17).
\newblock In {\em ICDAR}, volume~1, pages 1429--1434. IEEE, 2017.

\bibitem{sun2018textnet}
Y.~Sun, C.~Zhang, Z.~Huang, J.~Liu, J.~Han, and E.~Ding.
\newblock Textnet: Irregular text reading from images with an end-to-end
  trainable network.
\newblock In {\em ACCV}, 2018.

\bibitem{tian2017wetext}
S.~Tian, S.~Lu, and C.~Li.
\newblock Wetext: Scene text detection under weak supervision.
\newblock In {\em ICCV}, 2017.

\bibitem{tian2016detecting}
Z.~Tian, W.~Huang, T.~He, P.~He, and Y.~Qiao.
\newblock Detecting text in natural image with connectionist text proposal
  network.
\newblock In {\em ECCV}, pages 56--72. Springer, 2016.

\bibitem{wu2017self}
Y.~Wu and P.~Natarajan.
\newblock Self-organized text detection with minimal post-processing via border
  learning.
\newblock In {\em ICCV}, 2017.

\bibitem{xue2018accurate}
C.~Xue, S.~Lu, and F.~Zhan.
\newblock Accurate scene text detection through border semantics awareness and
  bootstrapping.
\newblock In {\em ECCV}, pages 370--387. Springer, 2018.

\bibitem{yang2018inceptext}
Q.~Yang, M.~Cheng, W.~Zhou, Y.~Chen, M.~Qiu, and W.~Lin.
\newblock Inceptext: A new inception-text module with deformable psroi pooling
  for multi-oriented scene text detection.
\newblock {\em arXiv preprint arXiv:1805.01167}, 2018.

\bibitem{yao2012detecting}
C.~Yao, X.~Bai, W.~Liu, Y.~Ma, and Z.~Tu.
\newblock Detecting texts of arbitrary orientations in natural images.
\newblock In {\em CVPR}, pages 1083--1090. IEEE, 2012.

\bibitem{yuliang2017detecting}
L.~Yuliang, J.~Lianwen, Z.~Shuaitao, and Z.~Sheng.
\newblock Detecting curve text in the wild: New dataset and new solution.
\newblock {\em arXiv preprint arXiv:1712.02170}, 2017.

\bibitem{zhang2016multi}
Z.~Zhang, C.~Zhang, W.~Shen, C.~Yao, W.~Liu, and X.~Bai.
\newblock Multi-oriented text detection with fully convolutional networks.
\newblock In {\em CVPR}, pages 4159--4167, 2016.

\bibitem{zhong2018anchor}
Z.~Zhong, L.~Sun, and Q.~Huo.
\newblock An anchor-free region proposal network for faster r-cnn based text
  detection approaches.
\newblock {\em arXiv preprint arXiv:1804.09003}, 2018.

\bibitem{zhou2017east}
X.~Zhou, C.~Yao, H.~Wen, Y.~Wang, S.~Zhou, W.~He, and J.~Liang.
\newblock East: an efficient and accurate scene text detector.
\newblock In {\em CVPR}, pages 2642--2651, 2017.

\bibitem{zhu2018sliding}
Y.~Zhu and J.~Du.
\newblock Sliding line point regression for shape robust scene text detection.
\newblock {\em arXiv preprint arXiv:1801.09969}, 2018.

\end{thebibliography}
}

\end{document}